\pdfoutput=1

\documentclass[11pt]{article}

\usepackage[final]{acl}

\usepackage{times}
\usepackage{latexsym}

\usepackage[T1]{fontenc}

\usepackage[utf8]{inputenc}

\usepackage{microtype}

\usepackage{inconsolata}

\usepackage{graphicx}

\usepackage{amsmath}
\usepackage{multirow}
\usepackage{subcaption}

\usepackage[most]{tcolorbox}
\newtcolorbox{myquote}[1][]{%
    colback=black!3,
    colframe=black!3,
    notitle,
    sharp corners,
    borderline west={2pt}{0pt}{blue!80!black},
    enhanced,
    breakable,
}

\title{Enhancing Goal-oriented Proactive Dialogue Systems via Consistency Reflection and Correction}

\author{
  Didi Zhang${\thanks{Equal contribution}}$, 
  Yaxin Fan${\footnotemark[1]}$, 
  Peifeng Li${\thanks{Corresponding author}}$, 
  \and Qiaoming Zhu \\
  School of Computer Science and Technology, Soochow University, Suzhou, China \\
  \texttt{\{ddzhang2023, yxfansuda\}@stu.suda.edu.cn}, 
  \texttt{\{pfli, qmzhu\}@suda.edu.cn} 
}

\begin{document}
\maketitle
\begin{abstract} 
Goal-oriented proactive dialogue systems are designed to guide user conversations seamlessly towards specific objectives by planning a goal-oriented path. However, previous research has focused predominantly on optimizing these paths while neglecting the inconsistencies that may arise between generated responses and dialogue contexts, including user profiles, dialogue history, domain knowledge, and subgoals.
To address this issue, we introduce a model-agnostic two-stage Consistency Reflection and Correction (CRC) framework. Specifically, in the consistency reflection stage, the model is prompted to reflect on the discrepancies between generated responses and dialogue contexts, identifying inconsistencies and suggesting possible corrections. In the consistency correction stage, the model generates responses that are more consistent with the dialogue context based on these reflection results. We conducted experiments on various model architectures with different parameter sizes, including encoder-decoder models (BART, T5) and decoder-only models (GPT-2, DialoGPT, Phi3, Mistral and LLaMA3), and the experimental results on three datasets demonstrate that our CRC framework significantly improves the consistency between generated responses and dialogue contexts.
Our code is publicly available at: \url{https://github.com/zhyidi/CRC}.
\end{abstract}

\section{Introduction}
\label{introduction}
\begin{figure*}[t]
\centering
\includegraphics[width=1\textwidth]{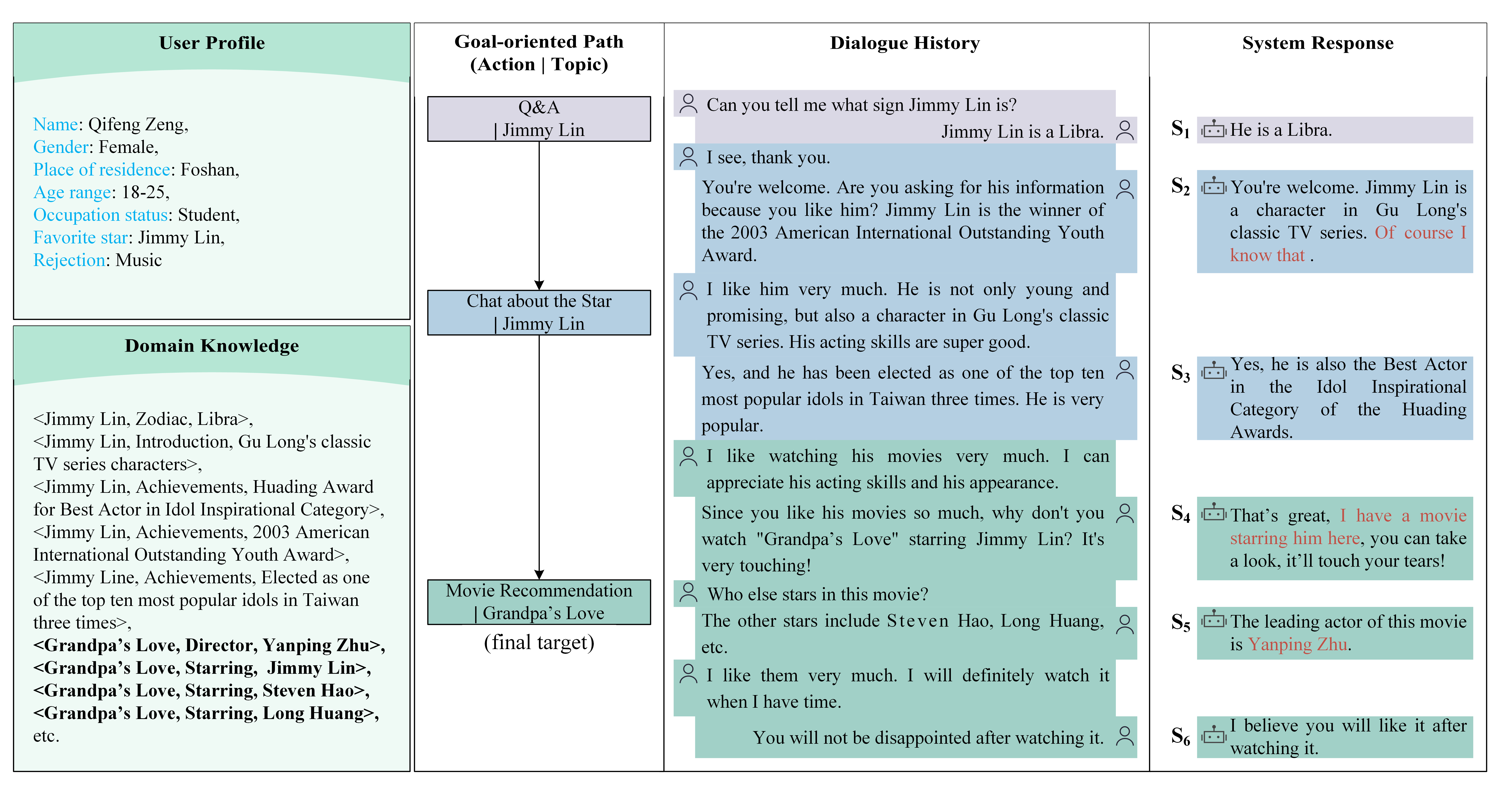} 
\vspace{-0.4cm}
\caption{An example of GPDS from the  DuRecDial \cite{liu-etal-2020-towards-conversational} dataset, where the system responses are generated by TPNet \cite{wang_TNNLS}.}
\label{fig_sample1}
\end{figure*}

The Goal-oriented Proactive Dialogue System (GPDS) focuses on achieving specific objectives by actively guiding and anticipating user needs \cite{liu-etal-2020-towards-conversational, wang_TNNLS, wang-etal-2023-target}. Unlike traditional dialogue systems that passively respond to user requests \citep{touvron2023llamaopenefficientfoundation, openai2024gpt4technicalreport}, GPDS strategically steers the conversation along a goal-oriented path, ensuring that a goal is naturally achieved while maintaining a positive user experience. GPDS has a wide range of applications in various domains, such as recommendation systems \cite{Fu_recommend, liu-etal-2020-towards-conversational, liu2021durecdial20bilingualparallel} and medical consultations \cite{xu2024reasoninglikedoctorimproving}. 


Figure~\ref{fig_sample1} presents an example of GPDS, which generates responses (e.g., $S_i$) based on the dialogue context, such as the user profile, dialogue history, domain knowledge, and subgoals within a goal-oriented path. GPDS can be divided into two primary sub-tasks: goal-oriented path planning and response generation. Initially, the system plans a goal-oriented path where each step is represented by an \texttt{<}action, topic\texttt{>} pair (e.g., ``Q\&A $\mid$ Jimmy Lin's constellation''$\rightarrow$ ``Chat about the Star $\mid$ Jimmy Lin''$\rightarrow$ ``Movie recommendation $\mid$ Grandpa's Love'' in Figure~\ref{fig_sample1}). Following this, the system generates responses aligned with the planned path, thereby guiding the conversation proactively and naturally toward achieving the final target (recommending the movie \emph{Grandpa's Love}).

Most previous studies on GPDS have primarily focused on planning goal-oriented paths using techniques such as CNN-based classifiers \cite{liu-etal-2020-towards-conversational}, Seq2seq paradigms \cite{Deng_MLF, wang-etal-2023-dialogue, Wang_Bidirectional, wang_TNNLS}, and graph interaction methods \cite{Zhang_Jia_Liu_Liu_Zhang_2024}. However, these approaches often overlook the inconsistencies that arise between generated responses and dialogue contexts. 
These inconsistencies manifest in several ways. First, there is an inconsistency with the dialogue history. As shown in Figure~\ref{fig_sample1}, the system asserts ``of course he knows Lin'' in $S_2$, even though the dialogue history does not inquire about his acquaintance with Lin. 
Second, there is an inconsistency with the subgoal. Although $S_4$'s action is to recommend a movie, it fails to address the topic of \emph{Grandpa's Love}, resulting in an invalid recommendation. 
Third, there is an inconsistency with the domain knowledge. $S_5$ states that Yanping Zhu is the star of \emph{Grandpa's Love}, whereas the domain knowledge indicates that Zhu is the director of the movie.
Lastly, there is an inconsistency with the user profile. The system might generate a response that does not align with the user's profile, as illustrated in Appendix~\ref{appendix_A}. For instance, the user's profile shows a preference for news about Nicholas Tse. However, the system recommends unrelated social news. Hence, these inconsistencies can lead to a poor user experience in real-world scenarios, causing conversations to abruptly break down and failing to achieve the intended targets. 

To address these inconsistencies, we propose a Consistency Reflection and Correction (CRC) framework, drawing inspiration from the reflective practice theory \cite{Checkoway1985TheRP} of human cognition that emphasizes the systematic reflection on experiences to identify and improve areas of weakness.
Specifically, in the \textbf{consistency reflection} stage, we guide the model to reflect its experience, i.e., whether its generated responses are consistent with the elements of dialogue contexts. If any inconsistencies are identified, the model is prompted to categorize the types of discrepancies and suggest potential corrections. In the \textbf{consistency correction} stage, the model is instructed to regenerate the responses more consistent with dialogue contexts based on the insights gained from the reflection stage. 

To validate the effectiveness of our framework, we conducted extensive experiments on three widely-used datasets: DuRecDial \cite{liu-etal-2020-towards-conversational}, DuRecDial 2.0 \cite{liu2021durecdial20bilingualparallel} and TopDial \cite{wang-etal-2023-target}. Since our framework is model-agnostic, we tested it on different model architectures and various parameter sizes, including encoder-decoder models (BART and T5) and decoder-only models (GPT-2, DialoGPT, Phi3-3.8B, Mistral-7B and LLaMA3-8B). The experimental results demonstrate that our CRC framework significantly improves the consistency between generated responses and dialogue context.

\section{Related Work}
\subsection{Goal-oriented Proactive Dialogue System}
Previous studies on GPDS typically began by planning a sequence of subgoals, followed by generating responses based on the subgoals to guide the conversation toward specific objectives. Most of them concentrated on goal-oriented path planning, employing the techniques such as CNN-base classifier \cite{liu-etal-2020-towards-conversational}, target-driven method \cite{wang2022followmeconversationplanning, 
wang_TNNLS}, Brownian bridge \cite{wang-etal-2023-dialogue}, prompt-based method \cite{Deng_MLF}, graph-grounded planning \cite{Liu_2023_tkde}, graph-interaction planning \cite{Zhang_Jia_Liu_Liu_Zhang_2024}, and bidirectional planning \cite{Wang_Bidirectional}.

However, these efforts often emphasize the importance of planning a goal-oriented path, overlooking the consistency between generated responses and dialogue contexts. This paper primarily focuses on response generation, aiming to improve GPDS by enhancing the consistency between generated responses and dialogue contexts.

\subsection{Discourse Consistency in Dialogue}
Discourse consistency in dialogue refers to the logical coherence and uniformity of information and themes, which is essential for fostering understanding and effective communication among participants. Previous research \cite{song-etal-2021-bob, 10191649, chen2023learning, zhou-etal-2023-simoap} has frequently employed Natural Language Inference (NLI) models to assess and enhance discourse consistency between generated responses and dialogue contexts. However, these NLI models often depend on additional training data, which can hinder their generalizability. In an innovative approach, we propose leveraging the model's inherent reflective capabilities to enhance the consistency between generated responses and dialogue contexts, thereby improving its generalizability.

\begin{figure*}[t!]
\centering
\includegraphics[width=1.01\textwidth]{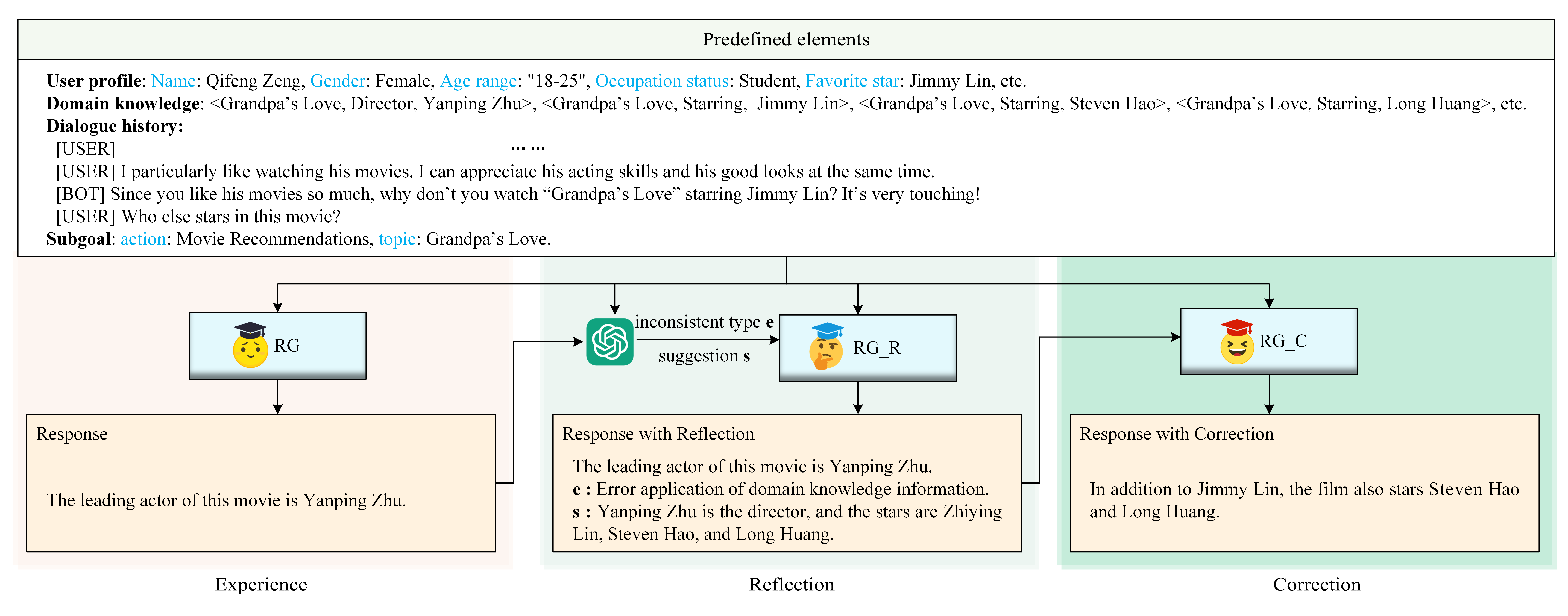} 
\vspace{-20pt}
\caption{Overview of our CRC framework.}
\label{fig_model}
\vspace{-5pt}
\end{figure*}

\section{Task Definition}
Given a dataset $D= \{U^i, K^i, H^i, G^i\}_{i=1}^N$, where $N$ is the size of the dataset. $U^i = \{u^i_{j}\}_{j=1}^{N_u}$ is the $i$-th user profile, where each item $u^i_{j}$ is a key-value pair, representing the user's personal information (e.g., name and gender). $K^i = \{k^i_{j}\}_{j=1}^{N_k}$ is the domain knowledge related to the $i$-th conversation of $D$, where each item $k^i_{j}$ is a triple \texttt{<}head, relation, tail\texttt{>}. $H^i = \{h^i_{m}\}_{m=1}^{M}$ is the content of the $i$-th conversation, consisting of $M$ turns. 
$G^i = \{g^i_{m}\}_{m=1}^M$ is the goal-oriented path for the dialogue $H^i$, where each $g^i_{m}$ consists of a dialogue action $a^i_{m}$ and a dialogue topic $t^i_{m}$. The final goal of the dialogue is represented by $g^i_M$. 
GPDS can be divided into two primary sub-tasks: goal-oriented path planning and response generation.

\vspace{0.1cm}
\noindent \textbf{Goal-oriented Path Planning}
Goal-oriented path planning aims to plan a sequence of subgoals to proactively guide the conversation to achieve the final target $g^i_{M}$. Each subgoal $g^i_{m} (1 \leq m \leq M)$ which is formulated as follows.
\begin{equation}
    g^i_{m} = \emph{GPP}(U^i, K^i,H^i_{\leq m} , G^i_{< m})
\end{equation}
where \emph{GPP} is a path prediction model, which has attracted the attention of most previous work.  In this paper, we adopted the same path prediction model as \citet{wang_TNNLS} and mainly focus on generating responses that are consistent with dialogue contexts. 

\vspace{0.1cm}
\noindent \textbf{Response Generation}
A generative model is employed to generate a response that aligns with the action and topic in $g^i_{m}$, thereby actively steering the conversation towards the final target. This process is represented as follows.

\begin{equation}
  r^i_m  = \emph{RG}(U^i, K^i, H^i_{\leq m}, g^i_{m})
\end{equation}
where $r^i_m$ is the generated response, and \emph{RG} denotes an autoregressive model.  Specifically, \emph{RG} autoregressively generates $r^i_m$ conditioned on the concatenation of the dialogue context, and it is optimized by minimizing the negative log-likelihood as follows.
\vspace{-0.1cm}
\begin{equation}
\begin{aligned}
\mathcal{L}(\theta) = -\sum_{i=1}^N \sum_{t=1}^T \log P(&r^i_{m, t} \mid r^i_{m, <t}, U^i, \\ 
& K^i, H^i_{\leq m}, g^i_{m}; \theta)
\end{aligned}
\end{equation}
where $\theta$ represents the trainable parameters, $r^i_{m, t}$ and $r^i_{m, <t}$ are the $t$-th token and the previous $t$-1 tokens of the response $r^i_m$, respectively, and $T$ is the length of $r^i_m$. In this paper, we focus on improving \emph{RG} to enhance the consistency between generated responses and  dialogue contexts.

\section{CRC Framework}
\textbf{Motivation}
Most of the Response Generation (RG) models in GPDS cannot produce responses that align with dialogue context. This is likely because their learning experience primarily involves imitation learning, which lacks deep reflection and correction mechanisms. According to the theory of reflective practice \cite{Checkoway1985TheRP}, the process of learning and growth comes from a cycle of experience, i.e., reflection and correction. Without this reflective practice, most models struggle to learn from their own experiences and summarize effectively, making it challenging for them to consistently generate responses that align with the dialogue context.  Therefore, establishing a framework that can involve the reflection and correction mechanism to improve the consistency between responses and dialogue contexts is crucial.

In this paper, we proposed a model-agnostic, two stage CRC framework including consistency reflection and consistency correction, as shown in Figure~\ref{fig_model}. In the consistency reflection stage, we first guide a \emph{RG} model to reflect on its experience, i.e., reflecting on the types of inconsistency between generated responses and dialogue contexts, and then suggest possible corrections. In the correction stage, we further guide the RG model to regenerate responses that are consistent with dialogue contexts on the reflection results.

\paragraph{Consistency Reflection}
As introduced in Section ~\ref{introduction}, the responses generated by the RG model may exhibit inconsistencies with dialogue context. These inconsistencies primarily pertain to the user profile \(U\), the domain knowledge \(K\), the dialogue history \(H\), and  the subgoal \(g\). To address this issue, we prompt the model to reflect on and consider ways to correct these inconsistencies. Specifically, we not only ask the RG model to generate responses, but also encourage it to analyze the types of inconsistencies between dialogue responses and dialogue contexts, providing suggestions for improvement. This can be formalized as follows.
\begin{equation}
    r^i_m, e^i_m, s^i_m = \emph{RG\_R}(U^i, K^i, H^i_{\leq m}, g^i_{m})
\end{equation}
where \emph{RG\_R} represents an autoregressive model with reflective ability, $e^i_m$ denotes the inconsistent type of the reflection results on the response $r^i_m$, and $s^i_m$ is a correction suggestion on the inconsistent type $e^i_m$. Therefore, the key lies in how to obtain the high-quality inconsistent type $e$ and the  correction suggestion $s$ to stimulate the model's reflective capabilities.

While manually annotating inconsistency types and correction suggestions is an ideal approach, the time-consuming and costly nature of manual annotation hinders its practical application. Thanks to the powerful understanding capabilities of Large Language Models (LLMs) like ChatGPT, which have already achieved success in data annotation across various fields \cite{wang-etal-2023-self-instruct, xu2023wizardlmempoweringlargelanguage}, we utilize ChatGPT \footnote{The version used is GPT-4o-2024-05-13.} to act as an annotator for the annotations of inconsistency types and correction suggestions. 
The prompt used is illustrated in Appendix~\ref{appendix_B}. 

We feed the dialogue context and the generated response to ChatGPT and let it evaluate the consistencies. ChatGPT first requires to identify the inconsistency types and then provide correction suggestions. An example is shown in Figure~\ref{fig_model}, where the inconsistency type is related to the domain knowledge, and the correction suggestion is that ``Yanping Zhu is the director, and the stars are Zhiying Lin, Steven Hao, and Long Huang.''. After obtaining the annotations $e$ and $s$, we continue to fine-tune the RG model to obtain a new RG\_R model with reflective capabilities. Let the concatenation of $r^i_m$, $e^i_m$, and $s^i_m$ be denoted as $c^i_m$, the optimization of RG\_R is as follows.
\begin{equation}
\label{eq5}
\begin{aligned}
    \mathcal{L}_{cr}(\theta) = -\sum_{i=1}^N \sum_{t=1}^T \log P(&c^i_{m, t} \mid c^i_{m, <t}, U^i, \\
    & K^i, H^i_{\leq m}, g^i_{m}; \theta)
\end{aligned}
\end{equation}
where $\theta$ represents the trainable parameters, $N$ is the data size and $T$ is the token length of $c^i_m$. During the learning process, the RG\_R model needs to generate not only the response \( r \) but also both the reflective results \( e \) and \( s \) regarding \( r \).

\vspace{0.1cm}
\noindent \textbf{Consistency Correction}
During the consistency correction phase, we continue to train RG to generate the response $ r_m^{i'} $ that is more consistent with the dialogue context on the reflective results from RG\_R. This can be formalized as follows.
\vspace{-0.1cm}
\begin{equation}
    r^{i'}_{m} = \emph{RG\_C}(U^i, K^i, H^i_{\leq m}, g^i_{m}, c^i_m)
\end{equation}
where \emph{RG\_C} represents an autoregressive model with correction ability.
Similar with Equ~\ref{eq5}, \emph{RG\_C} is trained by minimizing the negative log-likelihood as follows.

\begin{equation}
\label{r_loss_3}
	\begin{aligned}
	   \mathcal{L}_{cc}(\theta) = -\sum_{i=1}^N \sum_{t=1}^T \log &P(r^{i'}_{m, t} \mid r^{i'}_{m, <t}, U^i, \\ 
    & K^i, H^i_{\leq m}, g^i_{m}, c^i_m ; \theta)
	\end{aligned}
\end{equation}

\vspace{0.1cm}
\noindent \textbf{Training}
The training process is primarily divided into three stages. First, we train an initial model \emph{RG} by optimizing $\mathcal{L}$. Next, we enhance \emph{RG} by optimizing $\mathcal{L}_{cr}$ to obtain \emph{RG\_R}, which possesses reflective capabilities. Finally, we further optimize \emph{RG} by optimizing $\mathcal{L}_{cc}$ to achieve \emph{RG\_C}, which incorporates corrective capabilities.

\vspace{0.1cm}
\noindent \textbf{Inference}
During the inference phase, we first feed the dialogue context into \emph{RG\_R} to obtain \( c \), which includes the response \( r \), the inconsistency type \( e \), and the correction suggestion \( s \). Next, we feed both the dialogue context and \( c \) into \emph{RG\_C} to generate a response \( r' \) that is more consistent with the dialogue context.

\section{Experimentation}
\subsection{Experimental Settings}

\begin{table*}[t]
\centering
\resizebox{\textwidth}{!}{
\setlength{\tabcolsep}{3mm}
\begin{tabular}{lllllll}
\hline
\multicolumn{2}{c}{\textbf{Method}}                         & \multicolumn{1}{l}{\textbf{W F$_1$}} & \multicolumn{1}{l}{\textbf{BLEU-2}} & \multicolumn{1}{l}{\textbf{Dist-2}} & \multicolumn{1}{l}{\textbf{K F$_1$}} & \multicolumn{1}{l}{\textbf{Succ}} \\ \hline
\multirow{5}{*}{Previous methods} & MGCG         & 33.48    & 0.203    & 0.043    & 35.12    & 46.80    \\
                                    & UniMIND      & 40.58    & 0.231    & 0.078    & 44.51    & -    \\
                                    & TCP          & 41.40    & 0.299    & 0.072    & 48.63    &  68.57   \\
                                    & MGNN        & 43.50    & 0.274    & 0.064    & 45.00    & -    \\
                                    & GIGF  & 47.52    & 0.348    & 0.078    & 56.02    & -    \\
                                    \hline
\multirow{4}{*}{Encoder-Decoder}    & TP-BART\textsubscript{(140M)} & 37.22    & 0.255    & 0.083    & 44.52    & 71.50    \\
                                    & TP-BART w/ CRC       & 42.44$_{\uparrow5.22}$      & 0.280$_{\uparrow0.025}$   & 0.073    & 51.53$_{\uparrow 7.01}$    & 75.78 $_{\uparrow4.28}$    \\ \cline{3-7}
                                     & TP-T5\textsubscript{(390M)}           & 36.86    & 0.250    & 0.080    & 50.11    & 55.68    \\
                                    & TP-T5 w/ CRC       & 41.19$_{\uparrow4.33}$    & 0.278$_{\uparrow0.028}$    & 0.077   & 54.47$_{\uparrow4.36}$    & 72.25$_{\uparrow16.57}$    \\ \hline \cline{2-7} 
\multirow{8}{*}{Decoder-Only}      & TP-GPT2\textsubscript{(102M)} & 41.53    & 0.301    & 0.075    & 48.81    & 74.70    \\
                                    & TP-GPT2 w/ CRC       & 46.75$_{\uparrow5.22}$     & 0.344$_{\uparrow0.043 }$     & 0.074    & 54.77$_{\uparrow5.96 }$ &75.67$_{\uparrow 0.97}$      \\ \cline{3-7} 
                                    & TP-Dial\textsubscript{(99M)}        & 31.98    & 0.262    & 0.041    & 35.68    & 41.66    \\
                                    & TP-Dial w/ CRC & 43.76$_{\uparrow11.78}$     & 0.323$_{\uparrow0.061}$     & 0.062     & 52.66$_{\uparrow16.98}$     & 72.65$_{\uparrow30.99}$   \\ \cline{3-7} 
                                    & TP-Phi3\textsubscript{(3.8B)}        & 39.95    & 0.253    & 0.082    & 43.08    & 62.78    \\
                                    & TP-Phi3 w/ CRC & 44.67$_{\uparrow4.72}$     & 0.275$_{\uparrow0.022}$     & 0.082     & 49.18$_{\uparrow6.10}$     & 69.54$_{\uparrow6.76}$   \\ \cline{3-7} 
                                    & TP-Mistral\textsubscript{(7B)}        & 33.12    & 0.275    & 0.060    & 42.38    & 61.45    \\
                                    & TP-Mistral w/ CRC & 36.78$_{\uparrow3.66}$     & 0.311$_{\uparrow0.036}$     & 0.061     & 48.63$_{\uparrow6.25}$   & 71.92$_{\uparrow10.47}$     \\ \cline{3-7} 
                                    & TP-LLaMA3\textsubscript{(8B)}     & 40.24    & 0.276    & 0.095    & 51.11    & 60.72    \\
                                    & TP-LLaMA3 w/ CRC        & 45.96$_{\uparrow 5.72 }$    & 0.318$_{\uparrow0.042}$    & 0.095    & 56.86$_{\uparrow5.75}$    & 75.03$_{\uparrow14.31}$ \\ 
                                    & Golden-LLaMA3     & 43.57    & 0.301    & 0.099    & 53.43    & 65.64    \\
                                    & Golden-LLaMA3 w/ CRC     & 47.03$_{\uparrow 3.46 }$    & 0.317$_{\uparrow 0.016}$    &0.089     & 57.76$_{\uparrow 4.33}$    & 80.76$_{\uparrow15.12}$ \\
                                    \hline               
\end{tabular}
}
\caption{Experimental results on the Chinese DuRecDial dataset. The parameter sizes of the models are annotated as subscripts adjacent to the model names.}
\label{result_DuRecDial}
\end{table*} 
\begin{table*}[h!]
\centering
\resizebox{\textwidth}{!}{
\setlength{\tabcolsep}{3mm}
\begin{tabular}{lllllll}
\hline
\multicolumn{2}{c}{\textbf{Method}}                         & \multicolumn{1}{l}{\textbf{W F$_1$}} & \multicolumn{1}{l}{\textbf{BLEU-2}} & \multicolumn{1}{l}{\textbf{Dist-2}} & \multicolumn{1}{l}{\textbf{K F$_1$}} & \multicolumn{1}{l}{\textbf{Succ}} \\ \hline
\multirow{5}{*}{Previous methods} & MGCG         & 32.26    & 0.182    & 0.051    & 29.35    & 32.20    \\
                                    & UniMIND      & 33.66    & 0.184    & 0.135    & 23.75    & -    \\
                                    & TCP          & 33.12    & 0.201    & 0.070    & 34.86    & -    \\
                                    & MGNN        & 36.75    & 0.194    & 0.073    & 31.32    & -    \\
                                    & GIGF        & 38.35    & 0.241    & 0.089   & 66.91    & -    \\
                                    \hline
\multirow{4}{*}{Encoder-Decoder}    & TP-BART & 36.28    & 0.204    & 0.093    & 40.22    & 63.60    \\
                                    & TP-BART w/ CRC       & 38.87$_{\uparrow2.59}$    & 0.210$_{\uparrow0.006}$    & 0.094   & 60.62$_{\uparrow20.40}$    & 65.37$_{\uparrow1.77}$     \\ \cline{3-7} 
                                    & TP-T5           & 36.26    & 0.233    & 0.076    & 42.98    & 53.48    \\
                                    & TP-T5 w/ CRC       & 38.44$_{\uparrow2.18}$    & 0.229    & 0.076   & 63.58$_{\uparrow20.60}$    & 56.92$_{\uparrow3.44}$    \\ \hline
\multirow{8}{*}{Decoder-Only}       & TP-GPT2 & 34.62    & 0.217    & 0.082    & 38.80    & 60.70    \\ 
                                    & TP-GPT2 w/ CRC        & 36.02$_{\uparrow1.40}$    & 0.219$_{\uparrow0.002}$    & 0.083   & 59.73$_{\uparrow20.93}$    & 63.78$_{\uparrow3.08}$     \\\cline{3-7} 
                                    & TP-Dial    & 34.45    & 0.214    & 0.074    & 36.15    & 52.87    \\
                                    & TP-Dial w/ CRC        & 36.34$_{\uparrow1.89}$    & 0.215$_{\uparrow0.001}$    & 0.077  & 58.17$_{\uparrow22.02}$    & 55.05$_{\uparrow2.18}$    \\\cline{3-7} 
                                    & TP-Phi3        & 35.72    & 0.225    & 0.093    & 43.39    & 50.59    \\
                                    & TP-Phi3 w/ CRC & 38.56$_{\uparrow2.84}$     & 0.231$_{\uparrow0.006}$     & 0.092     & 57.45$_{\uparrow14.06}$     & 53.19$_{\uparrow2.60}$   \\ \cline{3-7} 
                                    & TP-Mistral        & 34.56    & 0.215    & 0.082    & 41.50    & 51.45    \\
                                    & TP-Mistral w/ CRC & 36.72$_{\uparrow2.16}$     & 0.228$_{\uparrow0.013}$     & 0.082     & 54.93$_{\uparrow13.43}$     & 53.12$_{\uparrow1.67}$   \\ \cline{3-7} 
                                    & TP-LLaMA3     & 36.00    & 0.230    & 0.095    & 43.72    & 51.85    \\
                                    & TP-LLaMA3 w/ CRC        & 38.54$_{\uparrow2.54}$    & 0.233$_{\uparrow0.003}$    & 0.096   & 65.91$_{\uparrow22.19}$    & 53.99$_{\uparrow2.14}$    \\
                                    & Golden-LLaMA3     & 37.02    & 0.231    & 0.096    & 46.19    & 52.92    \\
                                    & Golden-LLaMA3 w/ CRC        & 38.92$_{\uparrow1.90}$    & 0.233$_{\uparrow0.002}$    & 0.096   & 66.20$_{\uparrow20.01}$    & 56.37$_{\uparrow3.45}$    \\\hline
\end{tabular}
}
\caption{Experimental results on English DuRecDial 2.0.}
\label{result_DuRecDial2.0}
\vspace{-2pt}
\end{table*} 

\vspace{0.1cm}
\noindent \textbf{Datasets}
We conducted experiments on three widely recognized datasets: DuRecDial \cite{liu-etal-2020-towards-conversational}, DuRecDial 2.0 \cite{liu2021durecdial20bilingualparallel} and TopDial \cite{wang-etal-2023-target}. We followed the data processing procedures and splits outlined in previous work \cite{wang_TNNLS, Zhang_Jia_Liu_Liu_Zhang_2024} and the statistics are presented in Appendix~\ref{appendix_dataset_statistics}.

\vspace{0.1cm}
\noindent \textbf{Baselines}
We compared our CRC framework with several state-of-the-art baselines as follows. \textbf{MGCG} \cite{liu-etal-2020-towards-conversational} utilizes CNN for goal prediction and employs modified generation-based models for response generation. \textbf{UniMIND} \cite{Deng_MLF} unifies goal planning and response generation using prompt-based learning. \textbf{TCP} \cite{wang2022followmeconversationplanning} uses a Transformer-based planner to generate a sequence of actions and topic paths to guide response generation. \textbf{MGNN} \cite{Liu_2023_tkde} employs graph neural networks to model complex interactions between dialogue elements. \textbf{GIGF} \cite{Zhang_Jia_Liu_Liu_Zhang_2024} utilizes a directed heterogeneous graph to capture goal sequence information across different levels. TPNet \cite{wang_TNNLS} is an enhanced version of TCP that leverages several pre-trained models, including BART (denoted as \textbf{TP-BART}), GPT-2 (denoted as \textbf{TP-GPT2}), and DialoGPT (denoted as \textbf{TP-Dial}). In this paper, we adopted the same goal-oriented path as TPNet, primarily focusing on response generation. In addition to the aforementioned language models, we also applied our CRC framework  to T5 (denoted as \textbf{TP-T5}), 
Phi3 (denoted as \textbf{TP-Phi3}), Mistral (denoted as \textbf{TP-Mistral}) and LLaMA3 (denoted as \textbf{TP-LLaMA3}). Furthermore, we employed the golden goal-oriented path on LLaMA3 (denoted as \textbf{Golden-LLaMA3}) to demonstrate the general applicability of our CRC framework, independent of the performance of the goal-oriented path planning task.

\vspace{0.1cm}
\noindent \textbf{Evaluation Metrics}
We follow previous work \cite{wang_TNNLS} and use the following metrics: Word-level F$_1$ (W F$_1$), BLEU, Distinct (Dist), Knowledge F$_1$ (K F$_1$), and Goal Success Rate (Succ). Word-level F$_1$ measures the exact word overlap between generated and reference responses. BLEU measures the n-gram overlap with reference responses. Distinct evaluates the diversity of the generated responses. Knowledge F$_1$ measures the correctness of generated knowledge against domain knowledge triples. Goal Success Rate evaluates whether the dialogue successfully achieves both the target action and topic.

\vspace{0.1cm}
\noindent \textbf{Implementation Details} Please refer to Appendix~\ref{ImplementationDetails} for details. 

\subsection{Experimental Results}
\label{Experimental_Results}

The results of response generation on the three datasets are presented in Tables~\ref{result_DuRecDial} and~\ref{result_DuRecDial2.0}, as well as in Table \ref{tab:phi_mistral_results} in Appendix~\ref{Experimental_Results_on_different_datasets}. It can be observed that our CRC substantially enhances the performance of various model architectures across multiple metrics, showing notable improvements in Word-level F$_1$, BLEU-2, Knowledge F$_1$, and Goal Success Rate, while having minimal impact on Distinct. 
This suggests that by improving the consistency between generated responses and dialogue contexts, GPDS can more effectively guide conversations toward final targets without compromising the diversity of the responses. These findings demonstrate the effectiveness and generality of our CRC framework.

The observed enhancements in the Word-level $F_1$ and BLEU scores suggest that our framework enables the model to generate responses that more closely match the reference responses. The notable improvement in Knowledge $F_1$ can be attributed to our CRC framework, which prompts the model to address inconsistencies between responses and domain knowledge, thereby enhancing the model's ability to accurately utilize the domain knowledge. Likewise, the increase in Goal Success Rate is due to the fact that CRC can guide the model to identify and rectify discrepancies between responses and subgoals. By ensuring greater consistency between responses and each subgoal in the goal-oriented path, the model is better equipped to achieve the final objective. 

Comparing TP-LLaMA3 with other pre-trained models, TP-LLaMA3 has a natural advantage in terms of the Distinct and Knowledge $F_1$ metrics. 
This may be attributed to its incorporation of more diverse dialogue scenarios and knowledge-intensive corpora during the pre-training stage, which enhances its ability to generate diverse responses and accurately utilize domain knowledge.
Notably, our CRC can still significantly enhance the performance of TP-LLaMA3 across various metrics, indicating that our framework remains effective even for those LLMs with a larger number of parameters and can be effective supplement to LLMs.


Additionally, in comparison with TP-LLaMA3, Golden-LLaMA3 using annotated goal-oriented path improves all metrics on two datasets and these results indicate that the performance improvements in the task of goal-oriented path planning can enhance GPDS. Our CRC can further boost the performance of Golden-LLaMA3, demonstrating its generality, regardless of the performance of the goal-oriented path planning. Besides, the performance gaps between Golden-LLaMA3 and TP-LLaMA3 are relatively small, suggesting that further improvements in goal-oriented path planning may offer limited benefits for GPDS.

\begin{table}[t]
\centering
\resizebox{\columnwidth}{!}{
\begin{tabular}{llllll}
\hline
\textbf{Method} & \textbf{W F$_1$} & \textbf{BLEU-2} & \textbf{Dist-2} & \textbf{K F$_1$} & \textbf{Succ} \\ \hline
CRC   & 45.96 & 0.318 & 0.095 & 56.86 & 75.03 \\ 
w/o UP          & 43.48 & 0.306 & 0.095 & 56.30 & 73.00 \\  
w/o DH          & 43.62 & 0.311 & 0.093 & 55.73 & 73.49 \\ 
w/o DK          & 42.54 & 0.299 & 0.098 & 52.25 & 71.53 \\  
w/o SG         & 42.24 & 0.297 & 0.096 & 52.65 & 65.66 \\ \hline
\end{tabular}
}
\caption{Ablation results using TP-LLaMA3 on DuRecDial where UP, DH, DK and SG refer to user profile, dialogue history, domain knowledge and subgoal, respectively.}
\label{ablation_LLaMA3}
\end{table}

\section{Analysis}

\subsection{Ablation Study}

Our CRC aims to enhance response generation by improving consistency with the dialogue context, including the user profile, dialogue history, domain knowledge, and subgoals. Ablation experiments on the DuRecdial dataset using TP-LLaMA3 (see Table~\ref{ablation_LLaMA3}) reveal that removing any element leads to a performance decline in all metrics, highlighting the effectiveness of our CRC framework in maintaining and enhancing consistency with each element.

The removal of reflection and correction related to domain knowledge (w/o DK) significantly reduces Knowledge F1, highlighting the importance of consistency with domain knowledge for effective information utilization. Similarly, without subgoals (w/o SG), the Goal Success Rate drops markedly, demonstrating the importance of aligning responses with subgoals.
The absence of user profiles (w/o UP) and dialogue history (w/o DH) negatively impacts all metrics except Distinct, showing the benefits of maintaining consistency with user profiles and dialogue history in enhancing GPDS.

Additionally, regardless of which element is removed, the performance on the Distinct metric remains almost unchanged. This demonstrates that our CRC framework not only improves the model's ability to effectively guide conversations towards final targets but also maintains the model's capacity to generate diverse responses.

\begin{figure}[t]
\vspace{-15pt}
\centering
\includegraphics[width=1\columnwidth]{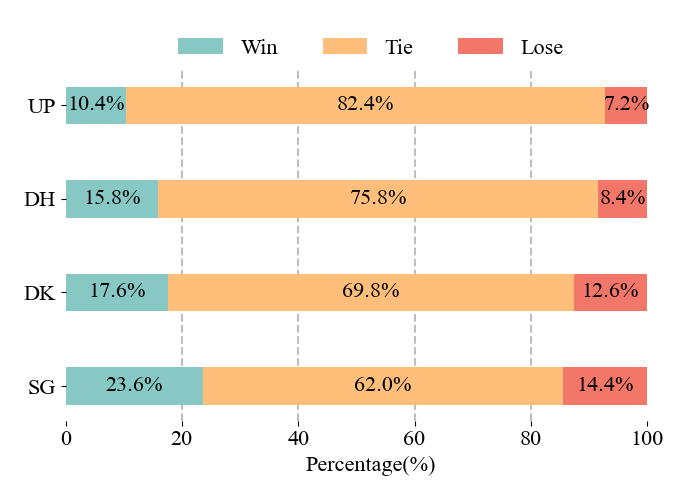}
\vspace{-20pt}
\caption{Pairwise evaluation results for TP-LLaMA3 w/ CRC vs. TP-LLaMA3 w/o CRC. }
\label{LLaMa3-8B_Consistency_analysis}
\end{figure}

\subsection{Consistency Analysis}
We conducted a pairwise human evaluation to compare the models with (w/) and without (w/o) CRC, assessing the consistency of the responses generated with each element of the dialogue context. We randomly selected 500 pairs of system responses from the DuRecDial dataset. 
The pairwise human evaluation results for TP-LLaMA3 are shown in Figure~\ref{LLaMa3-8B_Consistency_analysis}.  
The labels ``win'', ``tie'', and ``lose'' are used to indicate that TP-LLaMA3 w/ CRC is more consistent, equally consistent, or less consistent than TP-LLaMA3 w/o CRC, respectively. Appendix \ref{PairwiseHumanEvaluationResults} provides the details of human evaluation.

The tie rates for TP-LLaMA3 w/ CRC and w/o CRC decrease across user profile (UP), dialogue history (DH), domain knowledge (DK), and the subgoal (SG). This suggests that the challenge of generating responses consistent with these elements becomes more pronounced. 

Notably, TP-LLaMA3 w/ CRC exhibits a higher win rate compared to TP-LLaMA3 w/o CRC across all four elements. This underscores that our CRC framework effectively improves the consistency of the  TP-LLaMA3's responses and the elements of the dialogue context. Appendix \ref{PairwiseHumanEvaluationResults} provides the results of human evaluation using TP-BART, TP-T5, TP-GPT2 and TP-DialoGPT, which illustrate the same trend and further ensure the effectiveness of our CRC. 



\begin{figure}[t]
\vspace{-18pt}
\centering
\includegraphics[width=1\columnwidth]{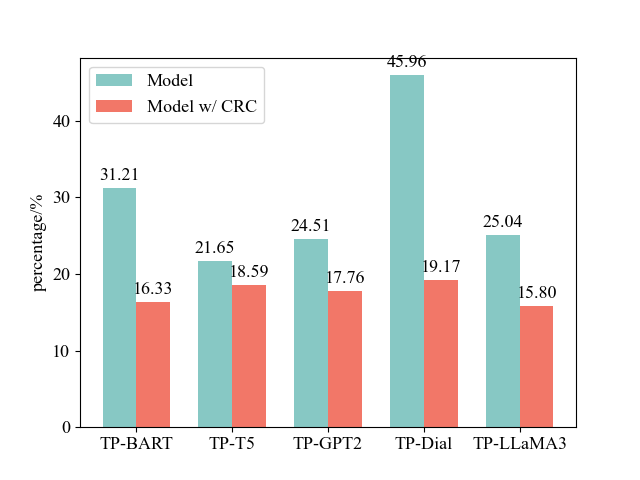}
\vspace{-24pt}
\caption{Subgoal failure rates on DuRecDial.}
\label{cn_Comparison_of_current_dialogue_goal_failure_rates}
\end{figure}

\subsection{SubGoals Failure Analysis}
It is essential for a goal-oriented proactive dialogue system to seamlessly steer the conversation towards the ultimate objective by generating the responses that align with each subgoal along the goal-oriented path. Consequently, we assessed whether each subgoal was successfully accomplished.
Figure~\ref{cn_Comparison_of_current_dialogue_goal_failure_rates} presents the subgoal failure rates (the rate at which the model are unable to achieve the subgoals) on DuRecDial for the  models both without and with CRC. 
It is evident that the models without CRC exhibit a higher rate of current turn goal failures, with a percentage exceeding 20\%. Such failures in subgoals have the potential to diminish the naturalness of the conversation, which may in turn lead to a poor user experience and complicate the achievement of the final objective. In contrast, our CRC framework has been demonstrated to significantly reduce the failure rate of subgoals, thereby facilitating a more natural conversational process. A similar trend is evident on the DuRecDial 2.0, as illustrated in Appendix~\ref{SubgoalsFailureonDuRecDial2_0Dataset}.

\subsection{Analysis of Inconsistency Detection and Reflection Content}
We provide an in-depth analysis of the ability of ChatGPT and our reflection model DialoGPT to detect inconsistencies and generate meaningful explanations. To evaluate the quality of reflections generated by both models, we randomly selected 500 samples each from the training and test sets and analyzed their performance in inconsistent type identification and the generation of accurate corrective suggestions, respectively.

The results show that ChatGPT correctly identified 94\% (245/261) of inconsistencies and generated accurate corrective suggestions for 97\% (237/245) of inconsistent cases (i.e., generating reasonable reflective content), demonstrating the high reliability of using ChatGPT for annotating reflection data.
In contrast, our reflection model correctly identified 94\% (227/242) of inconsistencies and provided accurate suggestions for 90\% (205/227) of the inconsistent cases. These results indicate that the reflection model successfully learned to identify and analyze inconsistencies from the annotations provided by ChatGPT, further validating the effectiveness of our CRC framework.

\begin{table*}[t]
\centering
\begin{tabular}{ccccccc}
\hline
\textbf{Reflection Model ($RG_R$)} & \textbf{Correction Model ($RG_C$)} & \textbf{W F$_1$} & \textbf{BLEU-2} & \textbf{Dist-2} & \textbf{K F$_1$} & \textbf{Succ} \\
\hline
- & - & 31.98 & 0.262 & 0.041 & 35.68 & 41.66 \\
DialoGPT & DialoGPT & 43.76 & 0.323 & 0.062 & 52.66 & 72.65 \\
DialoGPT & T5 & 41.02 & 0.265 & 0.062 & 54.20 & 72.15 \\
DialoGPT & BART & 41.72 & 0.266 & 0.074 & 50.87 & 75.55 \\
DialoGPT & LLaMA3 & 45.22 & 0.300 & 0.089 & 56.28 & 74.81 \\
\hline
\end{tabular}
\caption{Performance comparison on DuRecDial when the reflection model are DialoGPT and the correction model uses different model architectures.}
\label{performance_Combinations_1}
\end{table*}

\begin{table*}[t]
\centering
\begin{tabular}{ccccccc}
\hline
\textbf{Reflection Model ($RG_R$)} & \textbf{Correction Model ($RG_C$)} & \textbf{W F$_1$} & \textbf{BLEU-2} & \textbf{Dist-2} & \textbf{K F$_1$} & \textbf{Succ} \\
\hline
- & - & 31.98 & 0.262 & 0.041 & 35.68 & 41.66 \\
DialoGPT & DialoGPT & 43.76 & 0.323 & 0.062 & 52.66 & 72.65 \\
T5 & DialoGPT & 43.52 & 0.298 & 0.072 & 52.96 & 72.39 \\
BART & DialoGPT & 43.60 & 0.324 & 0.063 & 52.54 & 73.85 \\
LLaMA3 & DialoGPT & 44.23 & 0.328 & 0.065 & 53.48 & 73.62 \\
\hline
\end{tabular}
\caption{Performance comparison on DuRecDial when the correction model are DialoGPT and the reflection model uses different model architectures.}
\label{performance_Combinations_2}
\end{table*}

\begin{table}[t]
\centering
\resizebox{\columnwidth}{!}{
\setlength{\tabcolsep}{1.5 pt}
\begin{tabular}{ll}
\hline
\textbf{Type}    & \textbf{System Response}                                                                               \\ \hline
\multicolumn{2}{c}{\emph{\textbf{Consistency with the dialogue history}}}     \\                
w/o CRC & \begin{tabular}[l]{@{}l@{}}  You're welcome. Jimmy Lin is a character in Gu \\
Long's classic TV series. \textcolor{red}{Of course I know it.}  \end{tabular} \\ \cline{2-2}
w/ CRC  & \begin{tabular}[l]{@{}l@{}}  You're welcome. Jimmy Lin is a character in Gu \\
Long's classic TV series. \end{tabular}                         \\ \hline
\multicolumn{2}{c}{\emph{\textbf{Consistency with the domain knowledge}}}     \\  
w/o CRC & \begin{tabular}[l]{@{}l@{}}  The leading actor of this movie is \textcolor{red}{Yanping Zhu}. \end{tabular} \\ \cline{2-2}
w/ CRC  &  \begin{tabular}[l]{@{}l@{}} The leading actors in this movie, besides Jimmy \\  Lin, are \textcolor{blue}{Steven Hao} and \textcolor{blue}{Long Huang}. \end{tabular}                                                                                               \\ \hline
\multicolumn{2}{c}{\emph{\textbf{Consistency with the subgoal}}}     \\  

w/o CRC &        \begin{tabular}[l]{@{}l@{}} That’s great, I have a movie starring him here, \\ you can take a look, it’ll touch your tears! \end{tabular}  \\ \cline{2-2}
w/ CRC  &    \begin{tabular}[l]{@{}l@{}}  Have you seen the movie  \textcolor{blue}{\emph{Grandpa's Love}} \\starring him? I don't remember much about it, \\but I do remember the emotional bond between \\ the grandfather and grandson.   \end{tabular}  \\   \hline                                                        
\end{tabular}
}
\caption{A case generated by TP-GPT2.}
\label{cases}
\end{table}

\subsection{Analysis of Model Combination}

We examine the performance when the reflection model ($RG_R$) and the correction model ($RG_C$) differ in architecture, as detailed in table~\ref{performance_Combinations_1} and table~\ref{performance_Combinations_2}. 

The results in Table~\ref{performance_Combinations_1} show that incorporating any models as a correction model significantly enhances performance compared to a setup without CRC. However, the impact varies across different metrics. For instance, utilizing T5 as a correction model enhances K F$_1$ relative to DialoGPT; however, it does not demonstrate substantial benefits in other metrics. Conversely, BART demonstrates a marked enhancement in Dist-2 and Succ metrics. It is noteworthy that utilizing a more large model, such as LLaMA3, as the correction model, results in substantial enhancements across all metrics compared to DialoGPT, with the exception of BLEU-2. The BLEU score is calculated as the overlap between generated and reference responses, so increased diversity in the generated responses (as indicated by Dist-2) may have a negative effect on the BLEU-2 score. These findings imply that employing a correction model with a greater number of parameters can yield substantial performance enhancements.

Similarly, Table~\ref{performance_Combinations_2} demonstrates that using any framework model as a reflection model results in a substantial enhancement of performance in comparison to a model devoid of CRC. However, it is important to note that as the parameter size of the reflection model increases, its impact on the same correction model remains consistent.

In summary, the employment of a framework or parameter size for reflection or correction models has been shown to significantly enhance the initial model's performance. It is notable that a correction model with more parameters improves performance across most metrics. However, regardless of the architecture or parameter size of the reflection model, their contributions to model performance remain quite similar.

\subsection{Case Study}
We conducted case studies to demonstrate the effectiveness of our CRC framework in enhancing the consistency between generated responses and each element of the dialogue context , as shown in Table~\ref{cases} where the complete dialogue is presented in Figure~\ref{fig_sample1}.
Regarding the dialogue history,  TP-GPT2 emphasizes ``of course he knows Lin'' which reduces the consistency with the history. In contrast, our CRC improves the consistency by omitting this statement.
In terms of the domain knowledge,  TP-GPT2 incorrectly identifies the director of the movie as the leading actor. Conversely, our CRC correctly utilizes the domain knowledge (refer to the domain knowledge item in Figure~\ref{fig_sample1}).
For the subgoal, although TP-GPT2 mentions recommending a movie, it fails to address the topic of \emph{Grandpa's Love}, resulting in an ineffective recommendation. In contrast, our CRC successfully recommends the movie \emph{Grandpa's Love}. Additional case studies of TP-GPT2 concerning the user profile, as well as LLaMA3, are provided in Appendix~\ref{case_study}.

\section{Conclusion}
This paper introduces a model-agnostic Consistency Reflection and Correction framework CRC aimed at enhancing the consistency between the generated responses and the dialogue contexts in GPDS. The CRC framework adeptly guides models to detect and correct inconsistencies, thereby significantly boosting the performance. Comprehensive experiments and detailed analysis on the DuRecDial, DuRecDial 2.0 and TopDial datasets, across various model architectures and parameter scales, validate the effectiveness of our CRC framework. Future research will focus on maintaining consistency between generated responses and dialogue contexts while simultaneously enhancing the diversity of the generated responses.

\section*{Limitation}
Our Consistency Reflection and Correction (CRC) framework has two limitations. 
First, it relies on the closed-source GPT-4 as an annotator, raising concerns about the transparency and interpretability of the inconsistency detection process, despite our efforts to validate annotation quality. This dependence on a proprietary model may also limit the framework's applicability and reproducibility in settings with restricted access, even if the specific version of GPT-4 is disclosed. 
Second, while the CRC framework effectively improves consistency between generated responses and dialogue contexts, achieving a balance between this consistency and output diversity presents a challenge. Aiming for greater diversity may inadvertently introduce inconsistencies, complicating the framework's objectives and potentially reducing response quality. Addressing these limitations in future research will be crucial for optimizing the CRC framework and expanding its applicability across various dialogue systems.

\section*{Acknowledgments}
The authors would like to thank the three anonymous reviewers for their comments on this paper. This research was supported by the National Natural Science Foundation of China (Nos. 62376181 and 62276177), and Project Funded by the Priority Academic Program Development of Jiangsu Higher Education Institutions.

\bibliography{custom}

\newpage

\appendix

\begin{table*}[t]
\centering
\resizebox{\textwidth}{!}{
\begin{tabular}{p{3cm}p{13cm}}
\hline
User Profile         & \textbf{Name}:Xiaoling Liu; \textbf{Agreed News}: ``\textcolor{blue}{\textbf{Nicholas Tse} News}; Wang Leehom News'', etc.\\ \hline
Domain Knowledge             & \textless{}``Nicholas Tse'', News, ``Recently, Nicholas Tse opened a Weibo account, named `Fengwei', and certified as `Nicholas Tse, founder of Fengwei Holdings', which seems to be a big push to promote his own brand. On the evening of July 8, Nicholas Tse sent out his first Weibo: `Only here can you find the complete Nicholas Tse. Let's go together.'''\textgreater{}, etc. \\ \hline
Dialogue History          & \begin{tabular}[c]{@{}p{13cm}@{}}{[}USER{]} ``When is Nicholas Tse's birthday?''\\ {[}System{]} ``Nicholas Tse's birthday is August 29th.''\\ {[}USER{]} ``You are amazing, you know so much.''\\ {[}System{]} ``Thank you for your compliment, Nicholas Tse is also a male star with outstanding temperament in the entertainment industry, do you like him?''\\ {[}USER{]} ``I like Nicholas Tse very much.''\\ {[}System{]} ``I remember he also won the World Music Award for the best-selling singer in Asia, he is really excellent.''\\ {[}USER{]} ``Yes, yes, I hope he can get better and better.''\\ {[}System{]} ``Yes, by the way, I have some news about him here, do you want to see it?''\\ {[}USER{]} ``I am very interested, what is the news?''\end{tabular} \\ \hline
Subgoal & \textbf{Action}: News Recommendation; \textbf{Topic}: Nicholas Tse \\ \hline
TP-GPT2 w/o CRC        & ``Recent news reports said that in Hong Kong, \textcolor{red}{the police caught pickpockets, selling fake tickets, and impersonating staff to defraud money and tickets, and arrested more than a dozen suspects.}''  \\ 
TP-GPT2 w/ CRC        & ``Yes, there is. Recently, there was news that \textcolor{blue}{\textbf{Nicholas Tse} and his wife had a second child, but the second child is a daughter. Both children have grown up, but the second child is still very young and hasn't been born yet.}''  \\
\hline
\end{tabular}
}
\caption{Example of the inconsistency with the user profile.}
\label{sample_UP}
\end{table*}

\begin{table*}[t]
\centering
\resizebox{\textwidth}{!}{
\begin{tabular}{lccccc}
\hline
\textbf{Dataset} & \textbf{Language} & \textbf{Annotators} & \textbf{Domains} & \textbf{\#Dialogue} & \textbf{\#Utterance} \\ \hline
DuRecDial     & Chinese    & Crowd workers       & Movies, music, food, point-of-interest, restaurants & 8,004    & 126,186    \\
DuRecDial 2.0    & English    & Human experts       & Movies, music, food, point-of-interest, restaurants & 6,080     & 98,719     \\
TopDial       & English    & LLM agents          & Movies, music, food, point-of-interest, restaurants & 18,009    & 141,928    \\ \hline
\end{tabular}
}
\caption{Statistics of DuRecDial, DuRecDial 2.0 and TopDial.}
\label{Dataset_statistics}
\end{table*}

\begin{table*}[h!]
\centering
\setlength{\tabcolsep}{8pt}
\begin{tabular}{lccccccc}
\hline
\textbf{Parameter}    & \textbf{GPT2} & \textbf{BART} & \textbf{DialoGPT} & \textbf{T5} & \textbf{LLaMA3} & \textbf{Phi3} & \textbf{Mistral} \\ \hline
version  & base    & base    & base    & base    & 8B    & 3.8B    & 7B\\ 
learning rate  & 5e-5    & 2e-5    & 5e-5    & 1e-4    & 1e-4    & 1e-4    & 1e-4\\
batch size            & 8             & 8             & 8                 & 4           & 1       & 1       & 1    \\
msl         & 512           & 512           & 432               & 512         & 1024     & 1024      & 1024    \\
epoch            &10 &10 &10 &10 &10  &10  &10\\
mdl          &80 &80 &80 &80 &80  &80    &80        \\
optimizer         &Adam &Adam &Adam &Adam & AdamW  & AdamW  & AdamW \\       
decoding strategy          & \multicolumn{7}{c}{greedy search}  \\
\hline
\end{tabular}
\caption{Hyperparameter settings where ``msl'' denotes max sequence length and ``mdl'' denotes max decoding length.}
\label{Hyper-parameter}
\vspace{-2pt}
\end{table*}

\section{Example of Inconsistency with User Profile}
\label{appendix_A}
Table~\ref{sample_UP} shows an example of inconsistency with the user profile. The user profile indicates a preference for news about Nicholas Tse and Leehom Wang, yet TP-GPT2 recommends the social news from Hong Kong, which does not align with the user's stated interests. 

\section{Prompt for ChatGPT-based  Annotation}
\label{appendix_B}
we utilize ChatGPT \footnote{The version used is GPT-4o-2024-05-13.} to act as an annotator for the annotations of inconsistency types and correction suggestions. 
We feed the dialogue context and the generated response to ChatGPT and let it evaluate the consistency of them. The prompt is as follows: 
\begin{myquote}
Currently, the prediction task is performed: respond to user utterances based on information such as user profile, domain knowledge, dialogue history, and domain and current dialogue subgoal. However, there may be situations where the response is inconsistent with the four predefined elements above. For a dialogue, you need to analyze the AI Assistant's response from the perspective of whether the response is consistent with the four predefined elements above, and identifies inconsistency types and correction suggestions. The consistency requirements of the response with the four predefined elements are: \\ 
(1) Is user profile information applied? \\
(2) Is the consistency with the dialogue history maintained? \\
(3) Is domain knowledge information applied? \\ 
(4) Is the current dialogue subgoal achieved? \\ \\
{[}Start of Predefined Elements{]} \\ 
\$ \{User Profile\} \\
\$ \{Dialogue History\} \\
\$ \{Domain Knowledge\} \\
\$ \{Subgoal\} \\
{[}End of Predefined Elements{]} \\ \\
{[}Start of the Assistant's Response{]} \\
\$ \{Response\} \\
{[}End of the Assistant's Response{]} 
\end{myquote} 

\section{Statistics of Datasets}
\label{appendix_dataset_statistics}

THe DuRecDial and DuRecDial 2.0 datasets are the most widely used corpora, while the recently released TopDial dataset comprises a greater number of dialogues and a more diverse range of content, generated by LLM agents. The data tatistics of the three datasets are shown in Table~\ref{Dataset_statistics}.

\begin{table*}[h!]
\centering
\resizebox{\textwidth}{!}{
\begin{tabular}{p{2cm}p{7cm}p{7cm}}
\hline
\textbf{Phase} & \textbf{BART, T5, GPT2, DialoGPT} & \textbf{LLaMA3, Phi3, Mistral} \\ \hline
Experience & \begin{tabular}[l]{@{}p{7cm}@{}} \textbf{ Input}: {[}DK{]}\emph{delim}{[}SG{]}\emph{delim}{[}UP{]}\emph{delim}{[}DH{]} \\ \textbf{Output}: [Response]\end{tabular}  &\begin{tabular}[l]{@{}p{7cm}@{}}\textbf{Input}: Respond to user utterances based on domain knowledge, user profile, dialogue history, and the current dialogue goal. {[}DK{]}\emph{delim}{[}SG{]}\emph{delim}{[}UP{]}\emph{delim}{[}DH{]} \\ \textbf{Output}: [Response] \\ \end{tabular} \\ \hline
Reflection & \begin{tabular}[c]{@{}p{7cm}@{}}
\textbf{Input}: {[}DK{]}\emph{delim}{[}SG{]}\emph{delim}{[}UP{]}\emph{delim}{[}DH{]}\\ \emph{delim} \#\#\#stage2\_R\\ \textbf{Output}: {[}Response{]}\#\#\#{[}e{]}:\#\#\#{[}s{]}
\end{tabular} &\begin{tabular}[l]{@{}p{7cm}@{}}\textbf{Input}: Respond to user utterances based on domain knowledge, user profile, dialogue history, and the current dialogue goal, and annotate the response with the types of inconsistencies compared to predefined information, along with suggestions for generating better responses. {[}DK{]}\emph{delim}{[}SG{]}\emph{delim}{[}UP{]}\emph{delim}{[}DH{]} \\ \textbf{Output}: [Response]\#\#\#{[}e{]}:\#\#\#{[}s{]} \\ \end{tabular} \\ \hline
Correction & \begin{tabular}[c]{@{}p{7cm}@{}}
\textbf{Input}: {[}DK{]}\emph{delim}{[}SG{]}\emph{delim}{[}UP{]}\emph{delim}{[}DH{]} \\ \emph{delim} {[}Response{]}\#\#\#{[}e{]}:\#\#\#{[}s{]} \#\#\#stage3\_C\\ \textbf{Output}: {[}Corrected Response{]}\end{tabular} &\begin{tabular}[l]{@{}p{7cm}@{}}\textbf{Input}: Correct the pre-response and respond to user utterances based on domain knowledge, user profile, dialogue history, current dialogue goal, pre-response, types of inconsistencies between the pre-response and predefined information, and suggestions for generating a better response. {[}DK{]}\emph{delim}{[}SG{]}\emph{delim}{[}UP{]}\emph{delim}{[}DH{]} \\ \emph{delim}[Response]\#\#\#{[}e{]}:\#\#\#{[}s{]} \\ \textbf{Output}: {[}Corrected Response{]} \\ \end{tabular} \\ \hline
\end{tabular}
}
\caption{ The input and output formats for the experience, reflection, and correction stages. Delim is short for delimiter. The delimiters used by various models are shown in Table~\ref{delimiters}. The abbreviations used are: DK (domain knowledge), SG (subgoal), UP (user profile), DH (dialogue history), e (inconsistent type) and s (suggestion).}
\label{Prompt_splicing_details}
\end{table*}

\section{Implementation Details }
\label{ImplementationDetails}
Our implementation is mainly based on TPNet\footnote{https://github.com/iwangjian/Plan4RecDial}. We adopted the same goal-oriented path as TPNet, with a primary focus on response generation. The hyperparameters, detailed in Table~\ref{Hyper-parameter}, are kept consistent across the experience, reflection, and correction stages for all models. 

During training, we randomly selected 75\% of the training set for the experience stage, while the remaining 25\% was used for the reflection and correction stage. We employed LoRA \cite{hu2021loralowrankadaptationlarge} to fine-tune LLaMA3 \cite{touvron2023llamaopenefficientfoundation}, Phi3 \cite{abdin2024phi} and Mistral \cite{jiang2023mistral} using the LLaMA-Factory framework\footnote{\url{https://github.com/hiyouga/LLaMA-Factory}}.  The rank $r$ and scaling parameter $\alpha$ are set to 8 and 16, respectively.  The best model is selected based on its performance on the validation set. All our experiments were conducted on a single NVIDIA V100 GPU.

We follow previous work \cite{wang_TNNLS} by directly concatenating the elements of the dialogue context, using the default delimiters for different models to distinguish between these elements, as detailed in Table \ref{delimiters}. For LLaMA3, Phi3 and Mistral, we employ spaces to separate various types of elements. The input and output formats for the experience, reflection, and correction stages are presented in Table~\ref{Prompt_splicing_details}. For LLaMA3, Phi3 and Mistral, we provide appropriate prompts for each stage.

\begin{table}[t!]
\centering
\resizebox{\columnwidth}{!}{
\begin{tabular}{lcc}
\hline
\textbf{Model} & \textbf{DuRecDial}          & \textbf{DuRecDial2.0, TopDial} \\ \hline
GPT2\footnotemark           & {[}SEP{]}                   & \textless{}\texttt{|endoftext|}\textgreater{} \\
BART\footnotemark           & {[}SEP{]}                   & \textless{}/s\textgreater{} \\
DialoGPT\footnotemark       & {[}SEP{]}                   & \textless{}\texttt{|endoftext|}\textgreater{} \\
T5\footnotemark             & \textless{}/s\textgreater{} & \textless{}/s\textgreater{} \\
LLaMa3-8B\footnotemark      & Space                       & Space \\
Phi3-3.8B\footnotemark      & Space                       & Space \\
Mistral-7B\footnotemark      & Space                       & Space \\ \hline
\end{tabular}
}
\caption{Delimiters used by different models on three datasets.}
\label{delimiters}
\end{table}

\begin{table*}[t]
\centering
\setlength{\tabcolsep}{3mm}
\begin{tabular}{llllll}
\hline
\multicolumn{1}{l}{\textbf{Method}}   & \multicolumn{1}{l}{\textbf{W F$_1$}} & \multicolumn{1}{l}{\textbf{BLEU-2}} & \multicolumn{1}{l}{\textbf{Dist-2}} & \multicolumn{1}{l}{\textbf{K F$_1$}} & \multicolumn{1}{l}{\textbf{Succ}} \\ \hline
                        TP-Dial\textsubscript{(99M)}    & 33.87    & 0.162    & 0.068    & 36.62    & 43.88    \\
                        TP-Dial w/ CRC        & 36.12$_{\uparrow2.25}$     & 0.196$_{\uparrow0.034}$     & 0.069     & 42.51$_{\uparrow5.89}$     & 47.98$_{\uparrow4.10}$   \\ \cline{2-6} 
                        TP-Phi3\textsubscript{(3.8B)}        & 34.56    & 0.186    & 0.080    & 40.04    & 41.02    \\
                        TP-Phi3 w/ CRC & 37.12$_{\uparrow2.56}$     & 0.208$_{\uparrow0.022}$     & 0.079     & 44.23$_{\uparrow4.19}$     & 46.75$_{\uparrow5.73}$   \\ \cline{2-6} 
                        TP-Mistral\textsubscript{(7B)}        & 34.12    & 0.171    & 0.077    & 39.05    & 42.56    \\
                        TP-Mistral w/ CRC & 36.76$_{\uparrow2.64}$     & 0.192$_{\uparrow0.021}$     & 0.078     & 43.24$_{\uparrow4.19}$     & 46.67$_{\uparrow4.11}$   \\ \cline{2-6} 
                        TP-LLaMA3\textsubscript{(8B)}     & 35.94    & 0.182    & 0.082    & 40.77    & 43.32    \\
                        TP-LLaMA3 w/ CRC        & 37.45$_{\uparrow1.51}$     & 0.205$_{\uparrow0.023}$     & 0.082     & 47.89$_{\uparrow7.12}$     & 47.11$_{\uparrow3.79}$   \\ 
                        \hline               
\end{tabular}
\caption{Experimental results on TopDial. The parameter sizes of the models are annotated as subscripts adjacent to the model names.}
\label{tab:phi_mistral_results}
\end{table*}
\begin{figure*}[ht!]
  \centering
   \begin{subfigure}{0.45 \textwidth}
      \centering   
      \includegraphics[width=1\linewidth]{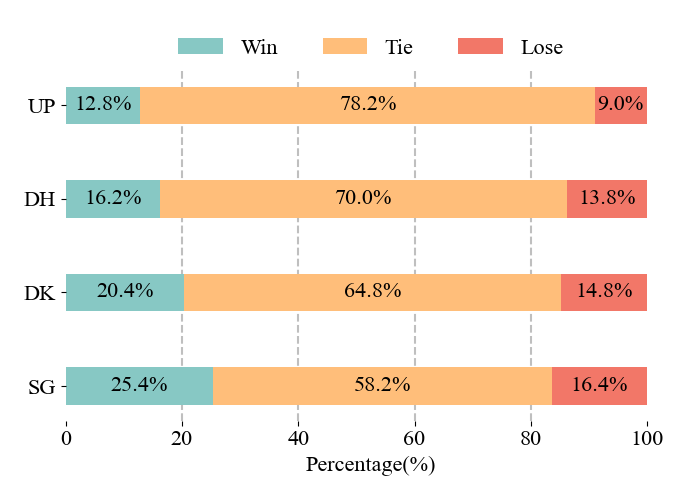}
        \caption{TP-BART w/ CRC vs. TP-BART}
        \label{Con:TPBART}
    \end{subfigure}
    \hfill
    \begin{subfigure}{0.45\textwidth}
      \centering   
      \includegraphics[width=1\linewidth]{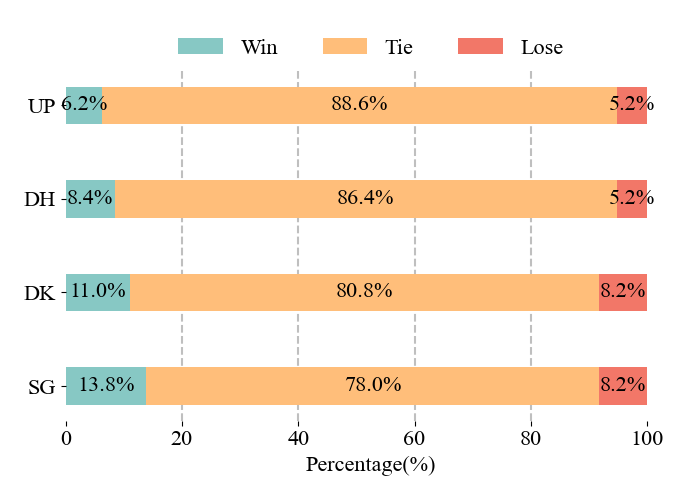}
        \caption{TP-T5 w/ CRC vs. TP-T5}
        \label{Con:TPT5}
    \end{subfigure}   
    \hfill
    \begin{subfigure}{0.45 \textwidth}
      \centering   
      \includegraphics[width=1\linewidth]{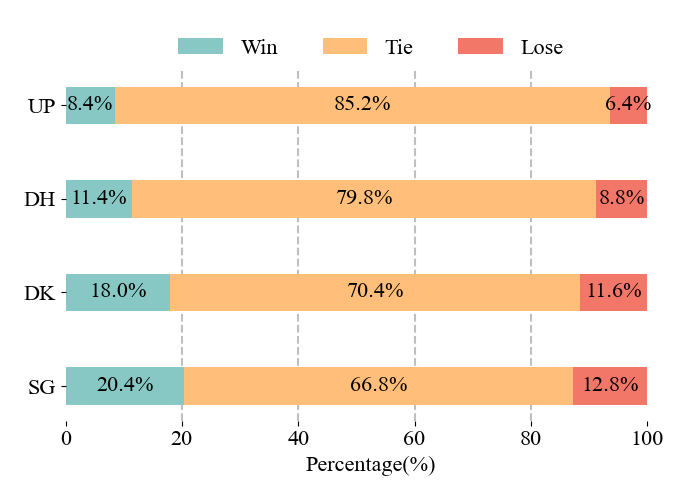}
        \caption{TP-GPT2 w/ CRC vs. TP-GPT2}
        \label{Con:TPGPT2}
    \end{subfigure}
    \hfill
    \begin{subfigure}{0.45 \textwidth}
      \centering   
      \includegraphics[width=1\linewidth]{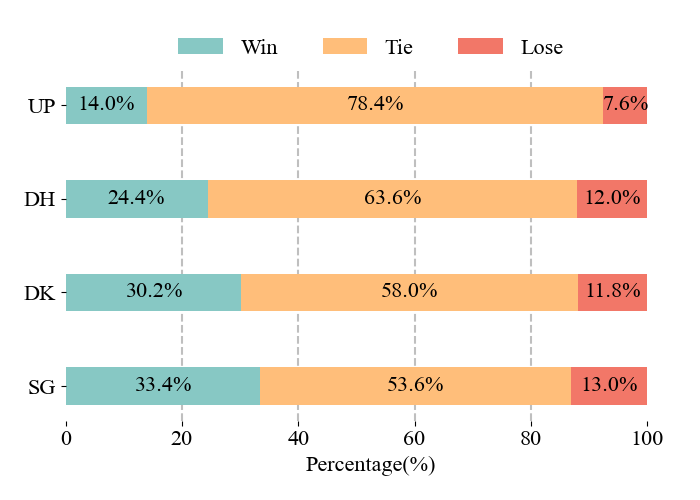}
        \caption{TP-Dial w/ CRC vs. TP-Dial}
        \label{Con:TPDialo}
    \end{subfigure}
\caption{ Pairwise evaluation results for TP-BART, TP-T5, TP-GPT2, and TP-Dial. Abbreviations: UP (user profile), DH (dialogue history), DK (domain knowledge), SG (subgoal). }
 \label{dm_proabability}
\end{figure*}

\section{Experimental Results on TopDial}
\label{Experimental_Results_on_different_datasets}

DuRecDial and DuRecDial 2.0 are the most widely used corpora in the GPDS field. To ensure a fair comparison with previous baselines, we mainly conducted experiments and analyses on these two corpora. Furthermore, the experiments are augmented by the incorporation of the recently released GPDS dataset, TopDial, which comprises a greater number of dialogues and a more diverse range of content, generated by LLM agents. The results, shown in Table~\ref{tab:phi_mistral_results}, indicate significant improvements across all evaluation metrics after integrating CRC, highlighting the effectiveness and generalizability of our CRC framework.

\section{Human Evaluation}
\label{PairwiseHumanEvaluationResults}

Assuming the system-generated responses before and after applying our method are defined as \(R_{\text{before}}\) and \(R_{\text{after}}\), respectively, we manually evaluated which of \(R_{\text{before}}\) and \(R_{\text{after}}\) is more consistent with the dialogue context, including user profiles, dialogue history, domain knowledge, and subgoals.

 During the manual evaluation, the evaluators are asked to assign one of three possible labels (i.e., ``win'', ``tie'', or ``lose'') to two anonymous responses, designated as A and B. One of these responses is \(R_{\text{before}}\), while the other is \(R_{\text{after}}\). The labels ``win'', ``tie'', and ``lose'' are used to indicate that A is more consistent, equally consistent, or less consistent than B, respectively.
Specific guidelines of the manual evaluation are as follows:

\textbf{Consistency with User Profile:} The evaluators first assess whether A and B contradict the user profile, such as gender, hobbies, age, etc., and then determine which response is more consistent. If both A and B are consistent or inconsistent with the user profile, the evaluators choose ``tie''.

\footnotetext[5]{\url{https://huggingface.co/uer/gpt2-chinese-cluecorpussmall/tree/main}}
\footnotetext[6]{\url{https://huggingface.co/fnlp/bart-base-chinese/tree/main}}
\footnotetext[7]{\url{https://huggingface.co/thu-coai/CDial-GPT_LCCC-base/tree/main}}
\footnotetext[8]{\url{https://huggingface.co/google/mt5-base}}
\footnotetext[9]{\url{https://huggingface.co/shenzhi-wang/Llama3-8B-Chinese-Chat/tree/main}}
\footnotetext[10]{\url{https://huggingface.co/microsoft/Phi-3-mini-4k-instruct}}
\footnotetext[11]{\url{https://huggingface.co/shenzhi-wang/Mistral-7B-v0.3-Chinese-Chat}}

\textbf{Consistency with Dialogue History:} The evaluators first assess whether  A and B accurately understand the user's intent and are logically coherent with the dialogue history, avoiding abrupt or irrelevant content. Then, the evaluators determine which response is more consistent with the dialogue history. If both A and B are consistent or inconsistent, the evaluators choose ``tie''.

\textbf{Consistency with Domain Knowledge:} The evaluators assess whether  A and B correctly utilize domain knowledge and choose the more consistent response. If both A and B correctly or incorrectly use domain knowledge, the evaluators choose ``tie''.

\textbf{Consistency with Subgoals:} The evaluators assess whether A and B correctly achieve the subgoals and choose the more consistent response. If both A and B achieve or fail to achieve the subgoals, the evaluators choose ``tie''.

We randomly selected 500 pairs of system responses and invited three NLP graduate students to independently evaluate them, taking the average of the results. The pairwise evaluation results for BART, T5, GPT2 and DialoGPT (Dial) are shown in Figure~\ref{dm_proabability}. It can be seen that all four models using CRC have a much higher win rate in all four aspects compared to these without CRC. The manual evaluation results also demonstrate the effectiveness of our CRC framework in improving consistency.




\begin{table*}[t]
\centering
\resizebox{\textwidth}{!}{
\begin{tabular}{p{3cm}p{13cm}}
\hline
User Profile         & \textbf{Name}:Liqian Niu; \textbf{Favorite Stars}: ``Leslie Cheung, Jiong He'', etc.\\ \hline
Domain Knowledge             & \textless{}``Jiong He'', Singing, ``Gardenia Blossoms''\textgreater{}, etc. \\ \hline
Dialogue History          & \begin{tabular}[c]{@{}p{13cm}@{}}{[}System{]} ``Good evening, Liqian Niu.''\\ {[}USER{]} ``Good evening!''\\ {[}System{]} ``How was work today?''\\ {[}USER{]} ``We got our bonuses today, so I'm very happy.''\end{tabular} \\ \hline
Subgoal & \textbf{Action}: \textcolor{blue}{\textbf{Music Recommendation}}; \textbf{Topic}: \textcolor{blue}{\textbf{Gardenia blooms}} \\ \hline
LLaMA3 w/o CRC        & ``Congratulations, \textcolor{red}{you can feel the fragrance of gardenia floating lightly in the air}.''  \\ 
LLaMA3 w/ CRC        & ``Congratulations, I recommend you a song `\textcolor{blue}{\textbf{Gardenia Blossoms}}'. The gentle music describes the mood of parting in a subtle way. The simple music rhythm makes people feel the fragrance of gardenias floating lightly in the air when listening.''    \\
\hline
\end{tabular}
}
\caption{Case study of the inconsistency with the subgoal.}
\label{sample_SG_llama3}
\end{table*}

\begin{table*}[h!]
\centering
\resizebox{\textwidth}{!}{
\begin{tabular}{p{3cm}p{13cm}}
\hline
User Profile         & \textbf{Name}:Shuaiguo Xu; \textbf{Favorite Stars}:Bo Huang, Leslie Cheung, \textbf{Age Range}:18-25, etc.\\ \hline
Domain Knowledge             & \textless{}``He's a Woman, She's a Man.'', Director, ``\textcolor{blue}{\textbf{Peter Chan}}''\textgreater{}, etc. \\ \hline
Dialogue History          & \begin{tabular}[c]{@{}p{13cm}@{}}{[}USER{]} ......,\\  {[}System{]} ``Then let's chat about the most gossipy celebrity topics. How about we talk about the lead singer of this song, Leslie Cheung?''\\ {[}USER{]} ``Leslie Cheung, huh? Not bad, not bad. He's an outstanding man, and his songs are really good.''\\ {[}System{]} ``Yeah, his songs are indeed quite good. He even won the Best Original Song award at the Golden Horse Awards in Taiwan.''\\ {[}USER{]} ``He's also the record holder for the best-selling Chinese album in South Korea. Impressive.''\\ {[}System{]} ``Indeed. By the way, he also starred in a very interesting movie, which has a bit of an autobiographical and metaphorical touch—it's called `Yang ± Yin: Gender in Chinese Cinema.' You should check it out.''\\ {[}USER{]} ``I actually really like that movie; I've already watched it.''\\ {[}System{]} ``You're quite a fan of his, then! In that case, let me recommend another one: `He's a Woman, She's a Man.' It explores love beyond gender. You should give it a try.''\\ {[}USER{]} ``I haven't seen that one yet, but who’s the director of this film?''\end{tabular} \\ \hline
Subgoal & \textbf{Action}: Movie Recommendation; \textbf{Topic}: He's a Woman, She's a Man. \\ \hline
LLaMA3 w/o CRC        & ``The director of this movie is \textcolor{red}{Leslie Cheung}.''  \\ 
LLaMA3 w/ CRC        &``The director is \textcolor{blue}{\textbf{Peter Chan}}.'' \\
\hline
\end{tabular}
}
\caption{Case study of the inconsistency with the domain knowledge.}
\label{sample_DK_llama3}
\end{table*}

\section{Results of Subgoals Failure}
\label{SubgoalsFailureonDuRecDial2_0Dataset}

\begin{figure}[h!]
\centering
\includegraphics[width=0.9\columnwidth]{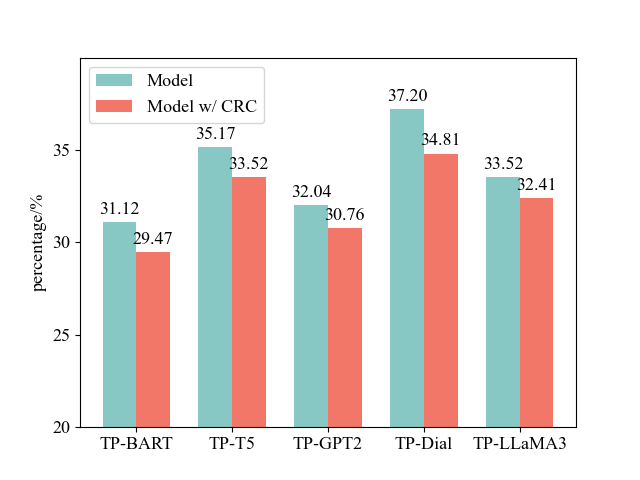}
\caption{Comparison of subgoal failure rates on DuRecDial2.0.}
\label{en_Comparison_of_current_dialogue_goal_failure_rates}
\end{figure}

As illustrated in Figure~\ref{en_Comparison_of_current_dialogue_goal_failure_rates}, our CRC  significantly reduces the goal failure rate for the current turn on the DuRecDial 2.0 dataset. This reduction not only further highlights the effectiveness of our framework but also demonstrates its robustness and adaptability across various dialogue scenarios. By decreasing the subgoal failure rates, our framework shows its potential to enhance user satisfaction through more reliable achievement of intended outcomes. Consequently, this improvement elevates the overall quality and utility of dialogue systems in practical applications.

\section{Case Study}
\label{case_study}

The case study of TP-GPT2 does not align with the user profile presented in Table~\ref{sample_UP}. TP-GPT2 recommends social news from Hong Kong, which does not match the user's stated interests. However, our TP-GPT2 with CRC can generate responses related to Nicholas Tse news, ensuring better consistency with the user profile.

Similarly, the case study of LLaMA3 does not align with the subgoal and domain knowledge illustrated in Tables~\ref{sample_SG_llama3} and~\ref{sample_DK_llama3}, respectively. Table~\ref{sample_SG_llama3} shows that LLaMA3 without CRC failed to recommend the song ``Gardenia Blossoms'' whereas LLaMA3 with CRC achieved this goal. Table~\ref{sample_DK_llama3} demonstrates that LLaMA3 without CRC failed to utilize domain knowledge, incorrectly identifying Leslie Cheung as the director of the movie \emph{He's a Woman, She's a Man}. In contrast, LLaMA3 with CRC correctly utilized the domain knowledge. Furthermore, LLaMA3 rarely exhibits inconsistencies with the user profile and dialogue history, thanks to its robust capabilities.

\end{document}